\documentclass[journal]{IEEEtran}
\usepackage{amsmath,amsfonts}
\usepackage{algorithmic}
\usepackage{algorithm}
\usepackage{array}
\usepackage[caption=false,font=normalsize,labelfont=sf,textfont=sf]{subfig}
\usepackage{textcomp}
\usepackage{stfloats}
\usepackage{url}
\usepackage{verbatim}
\usepackage{graphicx}
\usepackage{cite}
\usepackage{booktabs}
 \usepackage{color} 
\usepackage{makecell}
\usepackage{stfloats}
\usepackage[section]{placeins}
\usepackage{amssymb}
\usepackage{indentfirst}
\usepackage{subfig}
\DeclareUnicodeCharacter{E011}{\ensuremath{-}}
\usepackage[marginal]{footmisc}
\usepackage{multirow}
\begin{document}

\title{Federated Face Forgery Detection Learning with Personalized Representation}

\author{Decheng~Liu,~Zhan~Dang,~Chunlei~Peng,~\IEEEmembership{Member, IEEE}~,~Nannan~Wang,~\IEEEmembership{Senior Member, IEEE},~Ruimin Hu~and Xinbo Gao, \IEEEmembership{Fellow, IEEE}
\noindent
\thanks{D. Liu, Z. Dang, C. Peng and R. Hu are with the State Key Laboratory of Integrated Services Networks, School of Cyber Engineering, Xidian University, Xi’an 710071, Shaanxi, P. R. China (e-mail: dchliu@xidian.edu.cn; zd.xidian@gmail.com; clpeng@xidian.edu.cn; rmhu@xidian.edu.cn).\\
N. Wang is with the State Key Laboratory of Integrated Services Networks, School of Telecommunications Engineering, Xidian University, Xi’an 710071, Shaanxi, P. R. China (e-mail: nnwang@xidian.edu.cn).\\
X. Gao is with the Chongqing Key Laboratory of Image Cognition, Chongqing University of Posts and Telecommunications, Chongqing 400065, P. R. China.(e-mail: gaoxb@cqupt.edu.cn).}
}

\markboth{Journal of \LaTeX\ Class Files,~Vol.~14, No.~8, June~2024}%
{Shell \MakeLowercase{\textit{et al.}}: A Sample Article Using IEEEtran.cls for IEEE Journals}


\maketitle

\begin{abstract}
 Deep generator technology can produce high-quality fake videos that are indistinguishable, posing a serious social threat.
    Traditional forgery detection methods directly centralized training on data and lacked consideration of information sharing in non-public video data scenarios and data privacy. 
    Naturally, the federated learning strategy can be applied for privacy protection, which aggregates model parameters of clients but not original data.
    However, simple federated learning can't achieve satisfactory performance because of poor generalization capabilities for the real hybrid-domain forgery dataset.
   To solve the problem, the paper proposes a novel federated face forgery detection learning with personalized representation. 
   The designed Personalized Forgery Representation Learning aims to learn the personalized representation of each client to improve the detection performance of individual client models. 
   In addition, a personalized federated learning training strategy is utilized to update the parameters of the distributed detection model.
   Here collaborative training is conducted on multiple distributed client devices, and shared representations of these client models are uploaded to the server side for aggregation.
    Experiments on several public face forgery detection datasets demonstrate the superior performance of the proposed algorithm compared with state-of-the-art methods.
    The code is available at \emph{https://github.com/GANG370/PFR-Forgery.}
\end{abstract}

\begin{IEEEkeywords}
Facial forgery detection, personalized forgery representation learning, federated learning, privacy preserving.
\end{IEEEkeywords}

\section{Introduction}

\IEEEPARstart{W}{ith} the breakthrough of depth generation technology, face processing technology continues to emerge, resulting in high-quality fake videos that are indistinguishable from the naked eye. Therefore, they are easily abused by malicious users. For example, fake face images are used to deceive system authentication and fraud. Using synthetic fake face images for malicious intrusion, these phenomena are likely to cause serious social security problems. Therefore, detecting fake media content has become a hotspot of research. Although many deep fake detection methods have emerged, they are not effective for video data with large distribution differences in terms of complex sources, forgery methods, and personnel races. This brings great challenges to the face forgery detection task. Therefore, designing a more efficient face forgery detection model has become an important issue faced by the current biometric community and media forensics field.

\begin{figure}[t]
    \centering
    \includegraphics[width=0.48\textwidth]{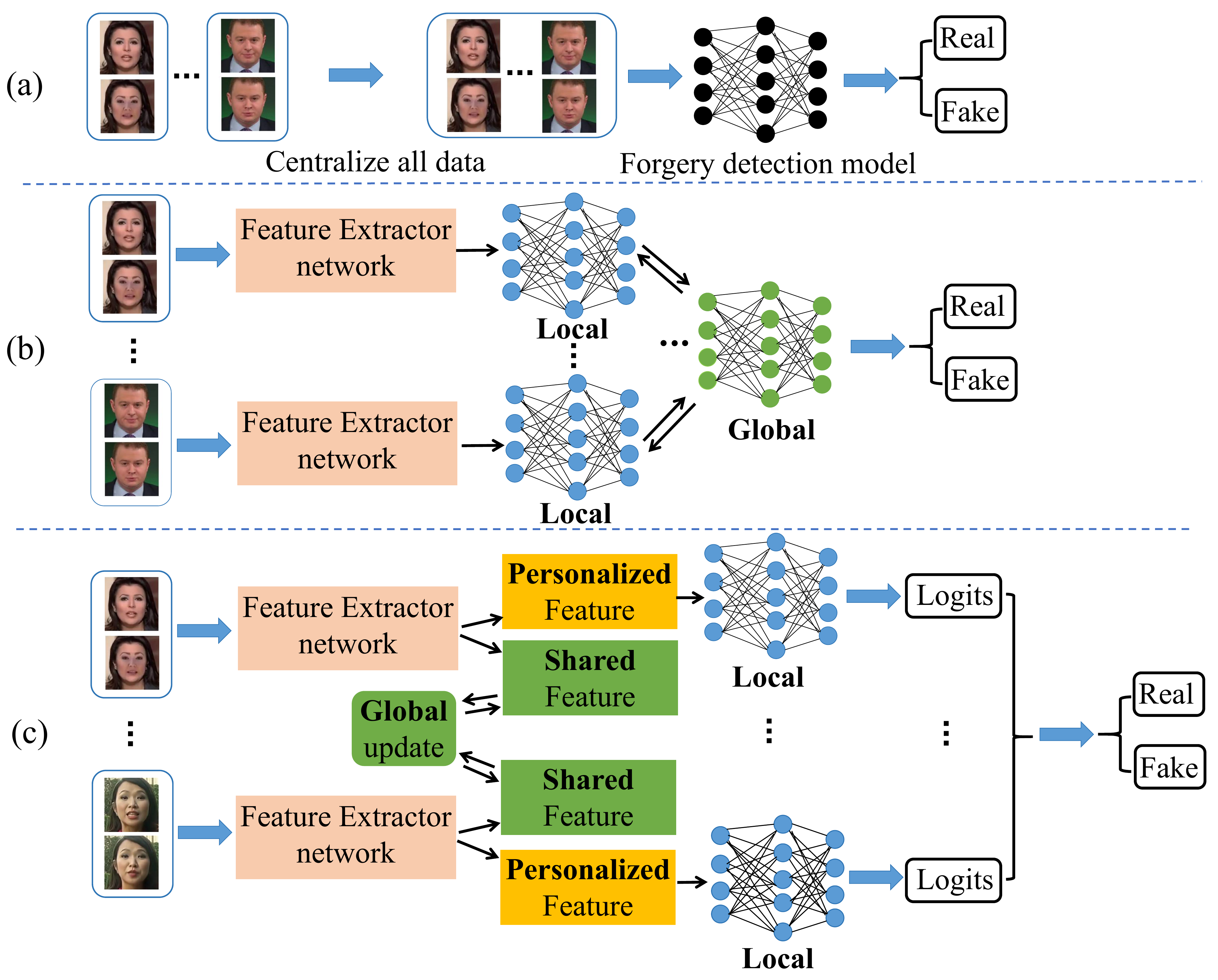}
    \caption{The differences between our method and traditional face forgery detection methods and traditional federated learning forgery detection method. (a) Traditional forgery detection methods require centralized aggregation of all client data for training, which is detrimental to privacy protection. (b) Traditional federated learning forgery detection method requires uploading all client model parameters to the server side, which prevents learning unique representations from individual clients. (c) Our method can extract personalized representations for complex forgery datasets with diverse types, and upload the shared representation to the server side for updates. In the testing stage, each client leverages its personalized model for local testing.}
\label{fig1}
\end{figure}

Existing face forgery detection can be roughly divided into two categories: image-based detection methods and video-based detection methods. Image-based detection methods mainly utilize image client artifacts \cite{chai2020makes}, mixed boundary information \cite{li2020face}, RGB images and their advanced semantic information in the frequency domain \cite{masi2020two}, etc. 
Different from forgery in image detection, video-based methods mainly use the inconsistency between forged video frames for authenticity detection.
Nowadays, the privacy of personal data draws more and more attention.
However, these mentioned traditional forgery detection algorithm lacks considering the privacy of training data.
As shown in Figure \ref{fig1} (a), 
traditional face forgery detection directly concentrates all client data for training, which can easily lead to the leakage of personal information and have serious consequences. 
Naturally, federated learning is a distributed and privacy-preserving machine learning strategy, which is explored in the forgery detection \cite{liu2023fedforgery} as shown in Figure \ref{fig1} (b).
However, related experiment results prove that simple federated learning can't adapt to real forgery detection scenarios, where there exist different types of forgery clues.
Thus, it is necessary to design a specific federated learning strategy for real forgery detection scenarios.
The designed personalization federated learning is an improved federated learning method that aims to improve the personalized performance and adaptability of the model \cite{collins2021exploiting}. As shown in Figure \ref{fig1} (c), the key challenge is how to disentangle personalized features and shared features in the client when training.

To solve the mentioned problem, we propose a federated face forgery detection learning with personalized representation. 
The proposed method can not only train forgery detection models with distributed non-public data but also improve the detection ability of complex forgery data. 
First, considering the diversity of distribution in the forgery dataset with mixed types, personalized forgery representation learning is designed to learn personalized features for client-side forgery types. 
Secondly, a personalized federated training strategy is introduced to cyclically update the shared representation of each client under the aggregation of the server, while the personalized part of each client is updated locally based on private data. 
In this framework, model parameters trained on clients are shared and aggregated, while training data on individual clients remains private, thus protecting the security of the data.

The main contributions of our paper can be summarized as follows:

\begin{enumerate}
    \item To the best of our knowledge, we first explore a federated face forgery detection learning with personalized representation, which further explores more robust face forgery clues by combining the shared representation of multiple distributed client models.
    
    \item The designed personalized forgery representation learning framework can disentangle shared features and personalized features. Personalized features contain more suitable properties of the client data set, further improving the applicability of complex forgery datasets with diverse types.

    \item Experimental results on public face forgery detection datasets show that the proposed algorithm has superior performance compared with state-of-the-art face forgery detection methods. The code is available at \emph{https://github.com/GANG370/PFR-Forgery.}
    
\end{enumerate}




\section{Related Work}
\subsection{Face Forgery Detection}
\label{A.Fake face detection}

Current detection methods can be divided into image-based forgery detection methods and video-based forgery detection methods.
Early-generation algorithms tend to produce obvious visual artifacts and inconsistencies between true and false directly in the facial area. Yang et al. \cite{yang2019exposing} determined the authenticity of the image by introducing inconsistencies in estimating 3D personalized poses from face images. Yang et al. \cite{yang2021learning} proposed a GAN fingerprint unwrapping network to unwrap fingerprint features from GAN-generated images. Luo et al. \cite{luo2021generalizing} used SRM technology to suppress the texture deviation of images, thereby avoiding the over-fitting problem of the training model. Dang et al. \cite{dang2020detection} proposed a mechanism to jointly predict binary labels of manipulated areas and attention maps to further improve forgery detection performance. The use of signal differences from the proposed face recognition network and background recognition network for image authenticity identification was proposed in \cite{nirkin2020deepfake}. Li et al. \cite{li2018exposing} found that forged images usually contain some distortion and blurring effects, thus proposing a more challenging face forgery detection method. Qian et al. \cite{qian2020thinking} applied DCT transform technology to images to collect frequency-aware clues, which can be used to further mine subtle artifacts in images. Chen et al. \cite{chen2020manipulated} proposed a multi-stream design and combined DFT features for image-level forgery detection. Cao et al. \cite{cao2022end} learned the common features of real faces by reconstructing face images and mined the essential differences between real faces and fake faces based on classification tasks to further improve detection performance. 
Luo et al. \cite{10315169} construct a fine-grained triplet and suppress specific forgery traces through data augmentation independent of prior knowledge. They design a progressive learning controller to guide the model to focus on the main feature components, allowing the network to learn key counterfeit features to achieve advanced counterfeit detection performance.
Miao et al. \cite{10004978} proposed a high-frequency fine-grained transformer network based on central differential attention and high-frequency wavelet sampler. Extensive experiments show that the specially designed framework performs well in the face of cross-datasets, cross-manipulations, and unseen perturbations. Face forgery detection is also very effective.
Tian et al. \cite{tian2024learning} proposed a categorical attention region proposal module that can localize false cues in the process.
Classification and supplementary learning modules to empower the network
Learn about richer false clues. The additional generated operation graphs can also serve as better supervision to enhance the performance of face forgery detectors.
Hua et al.\cite{hua2023learning}
convert the feature reconstruction layer into a deep neural network,
and at the same time, classification tasks and correspondence relationships will be optimized. The task is completed through alternative optimization. Therefore, the model can maintain high detection accuracy.
Yu et al.\cite{10173741} expressed the learning of the model as a meta-learning process and generated zero-shot face manipulation tasks for the model to learn meta-knowledge shared by diverse attacks. Experimental results show that the method achieves competitive results in the zero-sample setting.

Recent work treats deepfake detection as a video-level prediction problem and learns video-level features. Temporal features are often combined with spatial features for video forgery detection. Gu et al. \cite{gu2021spatiotemporal} proposed using a spatiotemporal discontinuity learning module to jointly learn the inconsistencies between single frames and consecutive frames in deepfake videos, respectively. Sun et al. \cite{sun2021improving} proposed temporal modeling based on precise geometric features to detect deepfake videos. Li et al. \cite{liy2018exposingaicreated} used the extracted eye region sequence through a convolutional neural network and lstm network to predict the blink probability to determine the authenticity of the video. With the further development of face forgery detection technology, the author \cite{ciftci2020fakecatcher} designed a video classifier based on physiological signal change synthesis to detect the authenticity of videos. Ganiyusufoglu et al. \cite{ganiyusufoglu2020spatio} used a three-dimensional convolutional neural network to model spatio-temporal features to capture the similarities between different deepfakes and further improve cross-domain fake video detection capabilities. However, these mentioned methods lack considering the privacy of training data.
Peng et al. \cite{peng2024deepfakes} observed the difference in gaze direction pattern distribution between real videos and fake videos, so they proposed to use a gaze analysis model to analyze the gaze features of face video frames and then applied a spatiotemporal feature aggregator to achieve gaze feature-based authenticity classification.
Wang et al. \cite{wang2023spatial} proposed a forgery cue enhancement network based on discrete cosine transform to achieve a more comprehensive representation of spectral-spatial and temporal features and make full use of the rich information in video sequences.
Ding et al. \cite{ding2024mining} proposed a spatiotemporal difference network to mine low-level clues for face forgery detection, and furthermore, used a multi-modal attention fusion module to effectively fuse complementary features from different branches.
Liu et al. \cite{liu2023hybrid} designed a hybrid spatiotemporal network that integrates spatial and temporal information in the same framework and jointly learns short-range and long-range relationships in the spatiotemporal dimension. A large number of experiments show that this method achieves excellent detection performance.
Zhao et al. \cite{zhao2023istvt} designed an interpretable spatiotemporal video transformer. The proposed decomposed spatiotemporal self-attention and self-reduction mechanisms are used to capture spatial artifacts and temporal inconsistencies to improve algorithm robustness.

\subsection{Federated Learning}

With the advancement of artificial intelligence technology, people's requirements for data privacy and security are becoming more and more stringent, thus promoting the development of the field of federated learning. Most existing federated learning training methods are derived from the federated averaging algorithm, which aims to train a well-performing global model. McMahan et al. \cite{mcmahan2017communication} proposed a federated learning algorithm for the first time, which can complete the weighted aggregation update of the model without direct contact with the training data. The optimization method of Yoshida et al. \cite{yoshida2020hybrid} is to make the server select client data to form a dataset that approximates IID. Sprague et al. \cite{sprague2018asynchronous} proposed a new asynchronous federated learning algorithm and applied it to image-based geolocation. Recently, W. Hongyi et al. \cite{wang2020federated} and others proposed a federated matching average algorithm, which uses coordinate weighted averaging and the arrangement invariance of network neurons to improve it. Liu et al. \cite{liu2023fedforgery} used residual federated learning and combined it with variational autoencoders to learn robust discriminative residual feature maps to detect face forgery cues. 
However, these mentioned methods have poor generalization capabilities for the complex forgery datasets with diverse types.
This paper proposes a personalized representation learning method. Each client improves the generalization of the algorithm by extracting personalized representations further exploring more robust face forgery clues of multiple distributed client models. 


\section{Proposed Approach}
\begin{figure*}[t]
    \centering
    \includegraphics[width=0.8\textwidth]{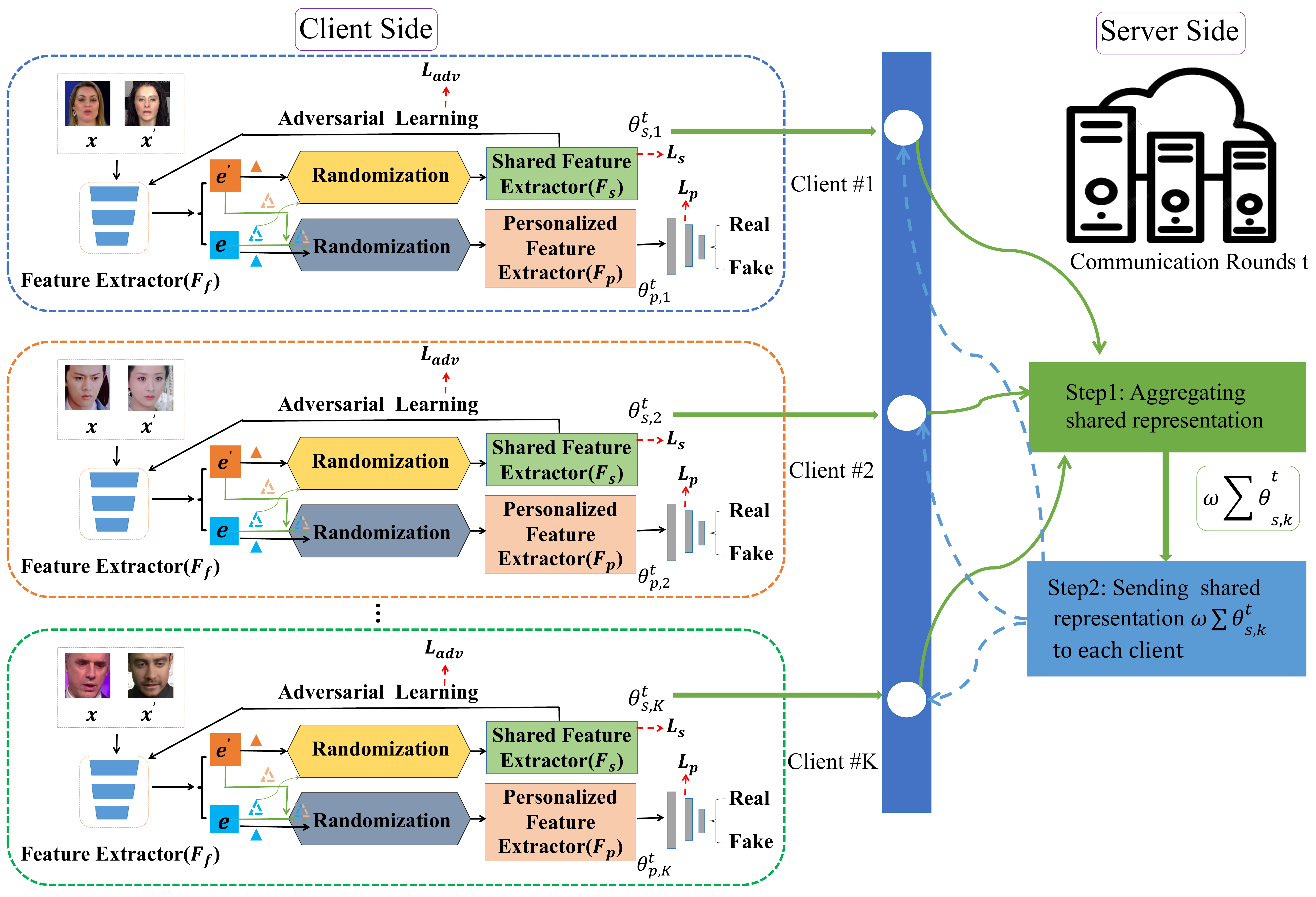}
    \caption{The overall framework of the proposed federated face forgery detection learning with personalized representation method.
}
\label{fig2}
\end{figure*}

This section details federated face forgery detection learning with personalized representation. The overall framework of the algorithm is shown in Figure \ref{fig2}. The designed personalized forgery representation learning framework can disentangle shared features and personalized features. Personalized features contain more suitable properties of the client data set, further improving the applicability of complex forgery datasets with diverse types. At the same time, by combining the shared representation of multiple distributed client models to explore more robust face forgery clues. 

\subsection{Motivation}
As shown in Figure \ref{fig1} (b), traditional federated learning directly uses the weighted average of all training parameters of each client as the basis for shared model parameter update, ignoring the inconsistency of data characteristics between different clients, resulting in poor performance of the model for real forgery detection scenarios. Some studies \cite{bengio2013representation} \cite{lecun2015deep} have shown that complex forgery datasets with diverse types may share a common representation. Hence, the paper proposes a novel federated face forgery detection learning with personalized representation. The designed personalized forgery representation learning framework can disentangle shared features and personalized features. Personalized features contain more suitable properties of the client data set, further improving the applicability of complex forgery datasets with diverse types. Aggregation updates are also performed on the server side by uploading the shared features of the client model, while each client's personalized features are retained locally for training updates, which further explores more robust face forgery clues. 
\emph{This not only protects the privacy of the client data sets but also further improves the generalization capabilities of complex forgery datasets with diverse types.}

\subsection{Preliminaries: Personalized Federated Learning}
Inspired by \cite{tan2022towards}, we consider using personalized federated learning methods when processing data from complex forgery datasets with diverse types. In order to further explore more robust face forgery clues, we design a novel federated face forgery detection learning with personalized representation. The personalized forgery representation learning framework can disentangle shared features and personalized features. Personalized features contain more suitable properties of the client data set, further improving the applicability of complex forgery datasets with diverse types. Additionally, the captured shared representation information further explores more robust face forgery clues by aggregating updates on the server side. As shown in Figure \ref{fig2}, the personalized forgery representation learning framework is designed to simulate data from complex sources in real scenarios. The specific settings are as follows: there are $K$ different clients, and the fake face dataset $D$ is distributed and stored in each client ($D$$_1$...$D$$_K$). Client data cannot be exchanged or transferred between clients. Clients train their own face forgery detection models locally. Since the data is a complex forged dataset with various types, traditional federated learning forgery detection directly concentrates all client parameters for training, which can easily lead to the inability to learn local personalized information and result in poor generalization ability. Therefore, uploading the shared representation of each client to the weighted average on the server side can not only ensure the privacy of the data but also improve the generalization ability of complex forged datasets with various types. Moreover, the designed personalized forgery representation learning framework can disentangle shared features and personalized features. Personalized features contain more suitable properties of the client data set, further improving the applicability of complex forgery datasets with diverse types. The personalized features of the $k$-th client is parameterized as $\theta$$_{p,k}$ ($k$ = 1, 2, 3, …, $K$). The shared features of the $k$-th client is parameterized as  $\theta$$_{s,k}$ ($k$ = 1, 2, 3, …, $K$).
In the testing stage, each client leverages its own personalized model for local testing.

\subsection{Personalized Forgery Representation Learning}

\begin{algorithm}[t]
    \caption{FedPR}
    \label{alg:algorithm}
    \textbf{Parameter}: Personalized personalized representation $\theta_{s}$ and shared representation  $\theta_{p}$; Number of communication rounds $t$; $w$ is the proportion of selected clients. $k$ represents the selected clients;
    \begin{algorithmic}[1] 
        \STATE Initialize $\theta_{s}^0$, $\theta_{p, 1}^0$, ..., $\theta_{p, K}^0$;
        \STATE Center Update:\\
        \FORALL{$t$=1, 2, ...,}
          \FORALL{selected client $k$=1, 2, ..., (in parallel)}
            \STATE Initializes $\theta_{p,k}^t$ $\leftarrow$ $\theta_{p,k}^{t-1}$
            \FORALL{$l$=1, 2, ...,}
              \STATE $\theta_{p, k}^{t, l+1}$ $\leftarrow$ ($L$$_k$ ($\theta_{p, k}^{t, l}$, $\theta_{s}^{t}$), $\theta_{p, k}^{t, l}$).
            \ENDFOR  
            \STATE Client $k$ locally updates the representation as:
\quad \quad \quad \quad $\theta_{s,k}^{t+1}$ $\leftarrow$ ($L$$_k$ ($\theta_{p,k}^{t}$, $\theta_{s}^{t}$), $\theta_{s}^{t}$)
           \STATE Client $k$ sends $\theta_{s,k}^{t+1}$ to server
         \ENDFOR
         \STATE  Server computes:
            $\theta_{s}^{t+1}$=$w$ $\sum$ $\theta_{s,k}^{t+1}$
        \ENDFOR
       
    \end{algorithmic}
\end{algorithm}

Considering that complex forged datasets with various types on the client will have a negative impact on traditional federated learning forgery detection. The key to personalized federated learning is to extract personalized representations of models trained by clients with complex forgery datasets with diverse types, as well as shared representation of multiple distributed client models. Therefore, we designed a novel federated face forgery detection learning with personalized representation. The framework of methods is shown in Figure \ref{fig2}. The client model consists of three parts: feature extractor, personalized feature extractor, and shared feature extractor. Input images $x$, $x'$, and the first pass through the feature extractor to obtain intermediate feature maps $e$, $e'$ $\in$ $R$$^{D\times{H}\times{W}}$. Then calculate the summary statistics (channel mean and standard deviation) and spatial configuration of $e$, $e'$. Adaptive instance normalization(AdaIN) \cite{huang2017arbitrary} is a method of instance normalization. Inspired by AdaIN  
 \cite{huang2017arbitrary}, we encourage network learning to pay more attention to features related to distinguishing true from false by interpolating feature statistics between different face samples.
 
\begin{equation}
R_{p}(e, \epsilon^*, \varphi^*)=\epsilon^*\cdot(\frac{e-\varphi(e)}{\epsilon(e)})+\varphi^*.
\end{equation}

Here $\varphi$(·) and $\epsilon$(·) represent the channel mean and standard deviation respectively, and $\varphi^*$, $\epsilon^*$ represent the linear interpolation results of the channel mean and standard deviation respectively.

For the intermediate feature maps $e$, $e'$ corresponding to the input image $x$ and the randomly selected image $x'$, we replace the spatial configuration of $e$ with the channel mean and standard deviation of $e'$ \cite{huang2017arbitrary}, 

\begin{equation}
R_s (e', \epsilon, \varphi)=\epsilon(e)\cdot(\frac{e'-\varphi(e')}{\epsilon(e')})+\varphi(e).
\end{equation}

The $R$$_{p}$ (e, $\epsilon^*$, $\varphi^*$) is fed into the personalized feature extraction network, and the feature extractor  F$_{f}$ and personalized feature extractor F$_{p}$ are further optimized through the loss function L$_{p}$:

\begin{equation}
L_{p}=-\sum_{n=1}^{N} y_nlog F_{p}[ R_{p} (e, \epsilon^*, \varphi^*) ]_n.
\end{equation}

By reorganizing the channel mean and standard deviation during the training process, the personalized forgery representation learning pays more attention to the personalized representation of the client training model when making decisions. Where $N$ is the number of class categories, $y$$\in${\{0,1\}}$^{N}$ is the one-hot label of input $x$.
The network is trained to learn shared features by minimizing the loss function $L$$_{s}$:
\begin{equation}
L_{s}=-\sum_{n=1}^{N}y_n log F_{s}[R_s (e', \epsilon, \varphi)]_n.
\end{equation}

In order to cooperate with personalized federated learning for information exchange to improve model generalization ability, we propose to use the shared feature extractor F$_{s}$ for adversarial learning. The feature extractor is trained with adversarial learning by minimizing the adversarial loss $L$$_{adv}$ calculated by the cross-entropy between the shared feature prediction and the uniform distribution.

\begin{equation}
L_{adv}=-\sum_{n=1}^{N}\frac{1}{N}logF_{s}[R_s(e', \epsilon, \varphi)]_n,
\end{equation}

\begin{equation}
L=\alpha L_{adv}+ \beta L_{p}+ \gamma L_{s}.
\end{equation}

Here $\alpha$ controls the weight of the adversarial learning part, $\beta$ controls the weight of personalized representation extraction, and $\gamma$ controls the weight of shared representation. The detailed procedures of the proposed FedPR algorithm are shown in Algorithm~\ref{alg:algorithm}.
In the testing stage, each client leverages its own personalized model for local testing.


\section{Experiments and Results}
\subsection{Dataset}
\begin{figure}[t]
 \centering
\includegraphics[width=0.48\textwidth]{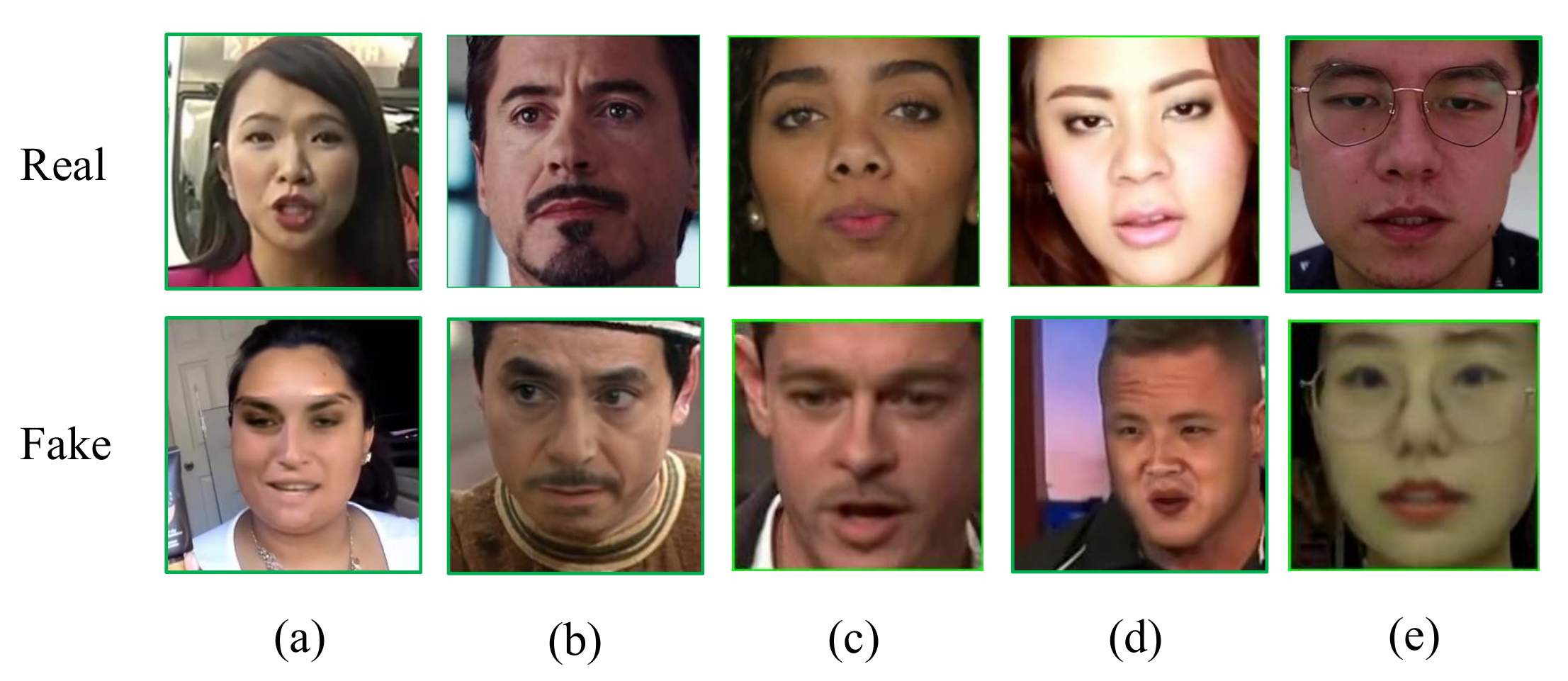}
\caption{The samples of public face forgery datasets: (a) FaceForensics++. (b)WildDeepfake. (c) CelebDF-v2. (d) Deeperforensics-1.0. (e) FMFCC-V.}
\label{fig3} 
\end{figure}

In this article, we use four public datasets, namely the FaceForensics++ dataset \cite{rossler2019faceforensics++}, WildDeepfake dataset \cite{zi2020wilddeepfake}, CelebDF-v2 dataset \cite{li2019celeb}, Deeperforensics-1.0 dataset \cite{jiang2020deeperforensics}, and FMFCC-V dataset \cite{li2022fmfcc}. Examples are shown in Figure \ref{fig3}. The first line displays real face samples, and the second line displays fake face samples. In addition, we also constructed the Forgery Source Hybrid Dataset to simulate complex real-life scenarios.

\textbf{FaceForensics++:} The real videos of the FaceForensics++ dataset are composed of 1000 real videos extracted from YouTube. Its fake videos include four subtypes: Deepfakes, Face2Face, FaceSwap and NeuralTextures. Each subset has different forgery methods and characteristics to simulate different forgery scenarios in the real world. In our personalized representation learning approach, the training data is divided into four age groups and assigned to eight customers, and the age range of every two customers is kept consistent.

\textbf{WildDeepfake:} The WildDeepfake dataset contains 7314 face sequences. The videos in WildDeepfake are collected from the Internet, and their face-changing videos are synthesized through various methods, which makes the detection of the WildDeepfake dataset more challenging. In our personalized-based federated learning method, the training data is divided into four races and assigned to eight clients, and the race types of each two clients are kept consistent.

\textbf{CelebDF-v2:} The CelebDF-v2 dataset contains 590 real original videos and 5639 corresponding fake videos collected from YouTube. Compared with the v1 version, this version of the dataset has excellent visual effects, so detection is more difficult. This dataset is currently widely used in the field of face forgery detection.

\textbf{Deepforensics-1.0:} The Deepforensics-1.0 dataset is a large-scale and widely used benchmark dataset for estimating face forgery detection methods. Fake videos are generated through an end-to-end face exchange framework. The authors also took into account factors such as different poses, lighting conditions, and expressions, so the dataset is more consistent with real-world scenarios. In our personalized-based federated learning method, the training data is first divided into five races and assigned to ten clients, and the race types of each two clients are kept consistent.

\textbf{FMFCC-V:} The FMFCC-V dataset is the first largest and public Asian DeepFake detection dataset. Among them, the real video sources come from 83 people, and the fake videos include four popular face-changing methods. In order to simulate real world scenarios, the authors introduced diversity to the deepfake video and the original video by adding 12 types of perturbations.

\textbf{Forgery Source Hybrid Dataset:} In order to effectively evaluate the face forgery detection performance of the proposed method in complex forgery datasets with diverse types, we used these four public datasets to construct a new mixed source dataset. The designed protocol details are as follows: four different types of FaceForensics++ dataset, WildDeepfake dataset, CelebDF-v2 dataset, and FMFCC-V dataset are used to form a mixed source dataset. The training set contains approximately 20,000 images of each subtype. The ratio of the training set to the test set is kept at 7:3. For the convenience of expression, we abbreviate Deepfakes data as FF++\_{DF}.

\subsection{Implementation Details}

For the video dataset, we extract 50 frames for each video. We choose the Dlib \cite{jeong2021eye} tool as the face detection extractor for the face forgery detection dataset. After the data is processed, only the facial area is cropped out as input to the face forgery detection model. We normalize all aligned faces and resize them to 256×256 before sending them to the network for training. Our personalized forgery representation learning network is implemented on the PyTorch platform. We adopt SGD with a momentum factor of 0.5 as the optimizer to minimize the loss function. The batch size is set to 32 and the initial learning rate is 0.01. In the following experiments, we set the weight coefficient $\alpha$ to 0.1, $\beta$ to 1 and $\gamma$ to 1 empirically.

\subsection{Comparison Results}

In this section, we compare the proposed method with state-of-the-art methods on constructed Forgery Source Hybrid Dataset and public datasets Deeperforensics-1.0, FaceForensics++, WildDeepfake and CelebDF-v2.
In order to fully evaluate the forgery detection performance, we use classification accuracy, the area under the receiver operating characteristic curve, and equal error rate as quantitative indicators. In the table, we use FedPR (w/o FL) to represent the results of training on centralized data, and FedPR (Ours) to represent the results of using personalized federated learning.
\begin{table*}[t]
	\centering
  \caption{The forgery detection accuracy (in \%) is evaluated by several state-of-the-art methods on four sub-datasets of the constructed forgery source hybrid dataset. The last column result is the average value. The best results are shown in black font, with the second place underlined.}
	\begin{tabular}{c|ccccc}
		\hline	
		Methods & CelebDF-v2 & FF++\_DF& FMFCC-V & WildDeepfake &Avg.		\\
		\hline
  ~&
  \multicolumn{5}{c}{\textbf{Without Considering Privacy}} \\
		\hline
	    CNNDetection$_1$ &64.98&94.18&88.06&71.36&79.65   \\
		Xception &72.12&87.54&75.54&57.77&73.27 \\
      	CNNDetection$_2$ & 65.79&\textbf{96.20}&83.99&63.61&77.40 \\
	 GFF &81.60&90.40&89.15&\underline{71.64}&83.20\\ 
   RFM &\underline{86.44}&94.77&99.22&65.92&86.59\\ 
   		CADDM &83.05&\underline{95.59}&74.69&\textbf{78.76}&83.02\\
FedPR (w/o FL) &84.29&94.12 &\underline{99.36}&70.05&\underline{86.96}\\
\hline
~&
\multicolumn{5}{c}{\textbf{Considering Privacy}} \\
\hline
FedPR (Ours)&\textbf{89.90}&94.81&\textbf{99.60}&70.81&\textbf{88.78}\\ 
	\hline
	\end{tabular}

 \label{tab:2} 
\end{table*}

\begin{table}[t]
	\centering
  \caption{Evaluation accuracy rate (in \%) and area under the receiver operating characteristic curve (in \%) of forgery detection performance on constructed Forgery Source Hybrid Dataset by several state-of-the-art methods. The best result is displayed in black font and the second place is underlined.}
	\begin{tabular}{c|cc}
		\hline	
		 Methods & Accuracy(\%) & AUC(\%)
		\\
		\hline
 ~&
\multicolumn{2}{c}{\textbf{Without Considering Privacy}} \\
\hline
		Xception&69.85 & 76.66 \\  
		CNNDetection$_{1}$ & 75.44  & 83.26 \\	
      	CNNDetection$_{2}$ & 78.57  & 87.48 \\
	    GFF & 82.71 & 91.18\\
            RFM & \underline{84.79} & \underline{92.76}\\
            CADDM & 81.51 & 90.40\\
\hline
~&
\multicolumn{2}{c}{\textbf{Considering Privacy}} \\
\hline
   FedPR (Ours) & \textbf{88.78} & \textbf{93.52}\\
	\hline
	\end{tabular}

 	\label{tab:1}  
\end{table}

\textbf{Results on Forgery Source Hybrid Dataset} In order to further prove the advantages of this proposed method, we trained and tested the models in the four subtypes included in the Forgery Source Hybrid Dataset. As shown in Table~\ref{tab:2}, our proposed method achieves the highest accuracy on the CelebDF-v2 and FMFCC-V datasets, outperforming 
 CADDM \cite{dong2023implicit},                 
   RFM \cite{wang2021representative}, 
    GFF \cite{luo2021generalizing} and other methods in most scenarios. This is because the designed personalized forgery representation learning can explore more robust face forgery clues. As shown in Table~\ref{tab:1}, we re-implemented several representative face forgery methods on the forgery source mixed dataset. Thanks to the designed personalized forgery representation learning, our proposed method achieves optimal performance in both auc and accuracy metrics. For example, compared with RFM \cite{wang2021representative}, the accuracy of our method exceeds 3.99\% and the auc exceeds 0.76\%. Compared with the existing CADDM method \cite{dong2023implicit}, this method also achieves competitive results. CNNDetection \cite{wang2020cnn} is susceptible to interference from specific generation methods. Therefore, two different data preprocessing methods are used to improve performance. CNNDetection$_{1}$ \cite{wang2020cnn} represents that the image may be gaussian blur or jpeged, each with 50\% probability, where gaussian blur parameters: $\sigma$\textasciitilde Uniform[0, 3], jpeged: the image is converted by two popular libraries OpenCV and PIL jpeg format, quality~uniform\{30, 31, …, 100\}; CNNDetection$_{2}$    \cite{wang2020cnn} represents that the image may be blurry and jpeg, 10\% probability. Our method improves the generalization ability for complex forgery datasets with diverse types by extracting personalized representations of clients and combining the shared representation of multiple distributed client models.

\begin{table}[t]
	\centering
	 \caption{Forgery detection performance evaluation was performed using several state-of-the-art methods on the Deepforensics-1.0 dataset, and the accuracy (in\%) was calculated. The best result is displayed in black font and the second place is underlined.}
	 
	\begin{tabular}{c|c}
		\hline	

		 Methods & Accuracy(\%) 
		\\
		\hline
 ~&
\multicolumn{1}{c}{\textbf{Without Considering Privacy}} \\
\hline
		C3D & 87.63  \\  
		TSN & 91.50  \\		
      	I3D & 90.75  \\
	   Resnet+LSTM  & 90.63\\
            Xception & 88.38 \\
           
             FedPR (w/o FL) & \textbf{98.64} \\
\hline
~&
\multicolumn{1}{c}{\textbf{Considering Privacy}} \\
\hline
   FedPR (Ours) & \underline{97.29} \\
	\hline
	\end{tabular}

 \label{tab:3} 
\end{table}

\textbf{Results on Deepforensics-1.0 dataset} As shown in Table~\ref{tab:3}, we compare the algorithm proposed in this paper with other representative face forgery detection methods. In order to prove the strong generalization ability, we selected 1000 manipulated videos in the standard set. When the training set is the standard set and the test set is single-level distortion, the proposed method can achieve good accuracy, even exceeding Resnet+LSTM 
 \cite{he2016deep} \cite{hochreiter1997long} 6.66\%, 
Xception \cite{chollet2017xception} 8.91\%, I3D 
   \cite{carreira2017quo} 6.54\%. We believe that these accuracy improvements are due to the designed personalized forgery representation learning. It is known that in some cases the accuracy of federated learning models is lower than that of centralized training models. In our experiments, the performance degradation caused by federated learning was 1.35\%.

\begin{table}[t]
	\centering
	 \caption{Evaluation AUC (in \%) and equal error rate (in \%) of forgery detection performance on WildDeepfake dataset by several state-of-the-art methods. The best result is displayed in black font and the second place is underlined.}
	
	\begin{tabular}{c|cc}
		\hline	       
		 Methods & AUC(\%) & EER(\%)
		\\
		\hline
 ~&
\multicolumn{2}{c}{\textbf{Without Considering Privacy}} \\
\hline
   Xception& 62.72 & - \\  
   RFM  & 57.75  & - \\ 
   ADD-Net & 62.35  & - \\
   F$^{3}$-Net & 57.10 & -\\
   MultiAtt &59.74 & -\\
   RECCE &64.31 & -\\
   LTW&67.12&39.22\\
   EN-B4 & 67.89 & \underline{37.21}\\
   GFF& 66.51 & 41.52\\
   SBI & 67.22 & 38.85\\
   FedPR (w/o FL) & \textbf{73.73} & \textbf{33.75}\\
\hline
~&
\multicolumn{2}{c}{\textbf{Considering Privacy}} \\
\hline
   FedPR (Ours) & \underline{68.25} & 37.84\\
	\hline
	\end{tabular}

 \label{tab:4}  
\end{table}

\textbf{Results on WildDeepfake dataset} As shown in Table~\ref{tab:4}, in order to further verify the generalization of the proposed method, we train on the FaceForensics++ dataset and test on the WildDeepfake dataset. Experimental results show that our proposed method can achieve good performance, even exceeding GFF  
 \cite{luo2021generalizing} 1.74\%, SBI  
  \cite{shiohara2022detecting} 1.03\%, LTW 
    \cite{sun2021domain} 1.13\%. The eer index of our method reaches 37.84\%, only lagging behind EN-B4 
      \cite{tan2019efficientnet} 0.63\%, while our method achieved auc of 73.73\% and eer of 33.75\% when using centralized training. It has reached the SOTA level on both indicators. This is due to the designed personalized features containing more suitable properties of the client data set, further improving the applicability of complex forgery datasets with diverse types.

\begin{table}[t]
	\centering
	 \caption{AUC (in \%) and equal error rate (in \%) of forgery detection performance on CelebDF-v2 dataset by several state-of-the-art methods. The best result is displayed in black font and the second place is underlined.}
	
	\begin{tabular}{c|cc}
		\hline	       
		 Methods & AUC(\%) & EER(\%)
		\\
		\hline
 ~&
\multicolumn{2}{c}{\textbf{Without Considering Privacy}} \\
\hline
   Capsule & 70.18 & - \\  
   Xception & 77.91  & 29.44 \\
   Add-Net & 62.12 & 41.51\\
   F$^{3}$-Net &  60.88 & 42.76\\
   Multi-Att & 76.95  & 28.11 \\
   PEL& 82.94 & \underline{24.24}\\
   FedPR (w/o FL) & \textbf{83.95} & 24.83\\
\hline
~&
\multicolumn{2}{c}{\textbf{Considering Privacy}} \\
\hline
   FedPR (Ours) & \underline{83.93} & \textbf{23.65}\\
	\hline
	\end{tabular}

 \label{tab:5}  
\end{table}
\textbf{Results on CelebDF-v2 dataset}  As shown in Table~\ref{tab:5}, in order to further verify the generalization ability of this method, we trained on the WildDeepfake dataset and tested on the CelebDF-v2 dataset. Experimental results show that our proposed method can achieve high auc, even exceeding 
Xception \cite{chollet2017xception} 6.02\%, Multi-Att \cite{nguyen2019multi} 6.98\%, PEL \cite{gu2022exploiting} 0.99\%. The eer index of our method reaches 23.65\%, which is better than PEL \cite{gu2022exploiting} 0.59\%. While our method uses centralized training, the auc reaches 83.95\% and the eer reaches 24.83\%. This is due to the designed personalized forgery representation learning, which can further explore more robust face forgery clues on complex forgery datasets with diverse types.

\subsection{Algorithm Analysis}
\begin{figure}[t]
 \centering
    \includegraphics[width=0.35\textwidth]{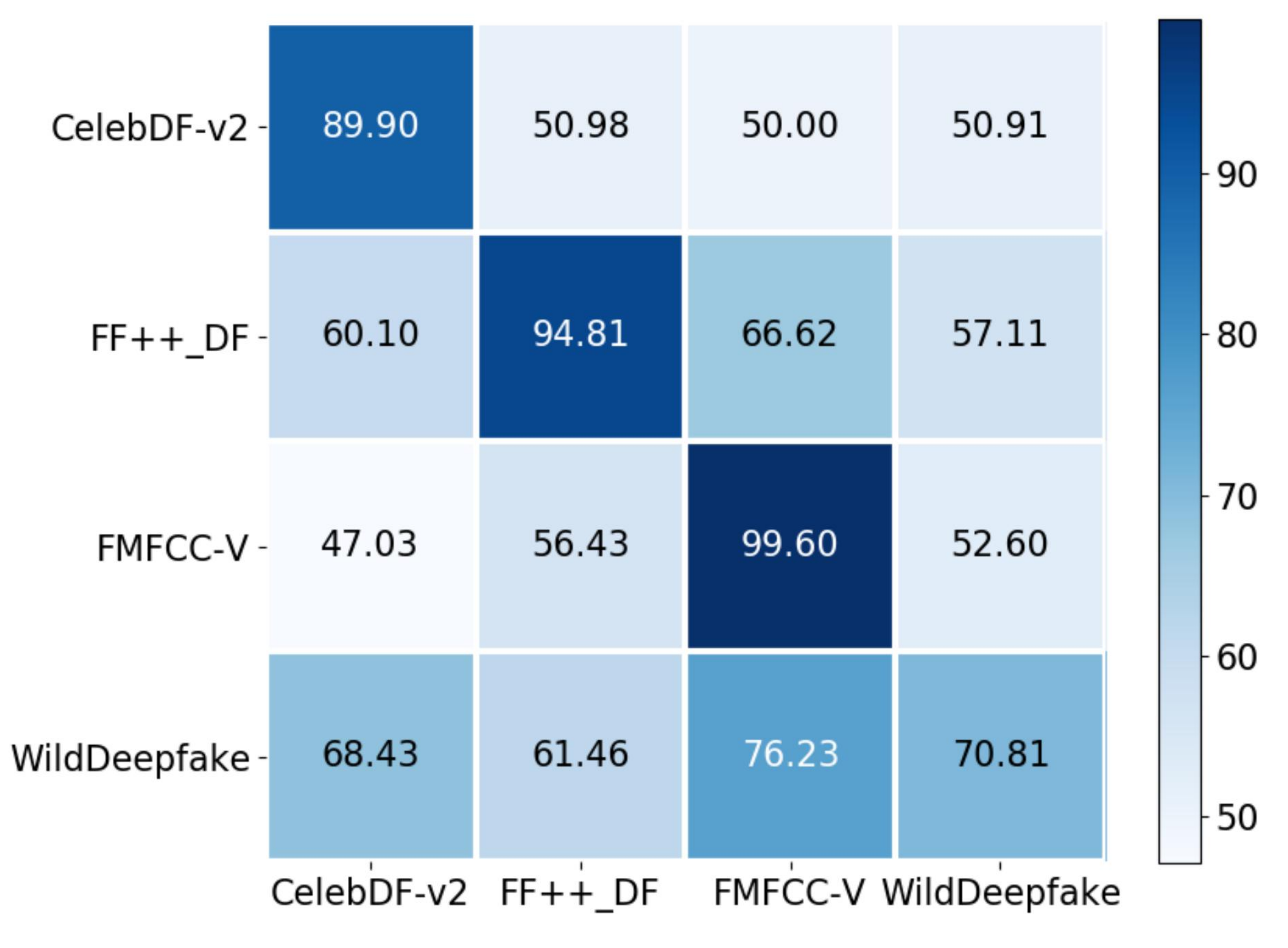}
\caption{Results of cross-validation on different training and testing subsets. The vertical axis represents training data, the horizontal axis represents testing data, and the evaluation index is accuracy. }
\label{fig4}
\end{figure}

\textbf{The effect of personalized features.}
To demonstrate that our proposed method can extract personalized representations of customers, we conducted cross-validation experiments on four subtypes included in the fake source hybrid dataset, and the results are shown in Figure \ref{fig4}. It can be seen from Figure \ref{fig4} that the color of the diagonal part is the darkest, and the rest of the color is lighter. For example, the personalized model is trained on the client holding the CelebDF-v2 data set, where the forgery detection accuracy of the local client can reach 89.90\% on the CelebDF-v2 data set.
But the detection accuracy for other clients' data FF++\_DF, FMFCC-V and WildDeepfake is poor. The same is true for other clients. This also proves that the personalized features contain more suitable properties of the client data set, further improving the applicability of complex forgery datasets with diverse types.

\begin{figure}[t]
 \centering
\includegraphics[width=0.4\textwidth]{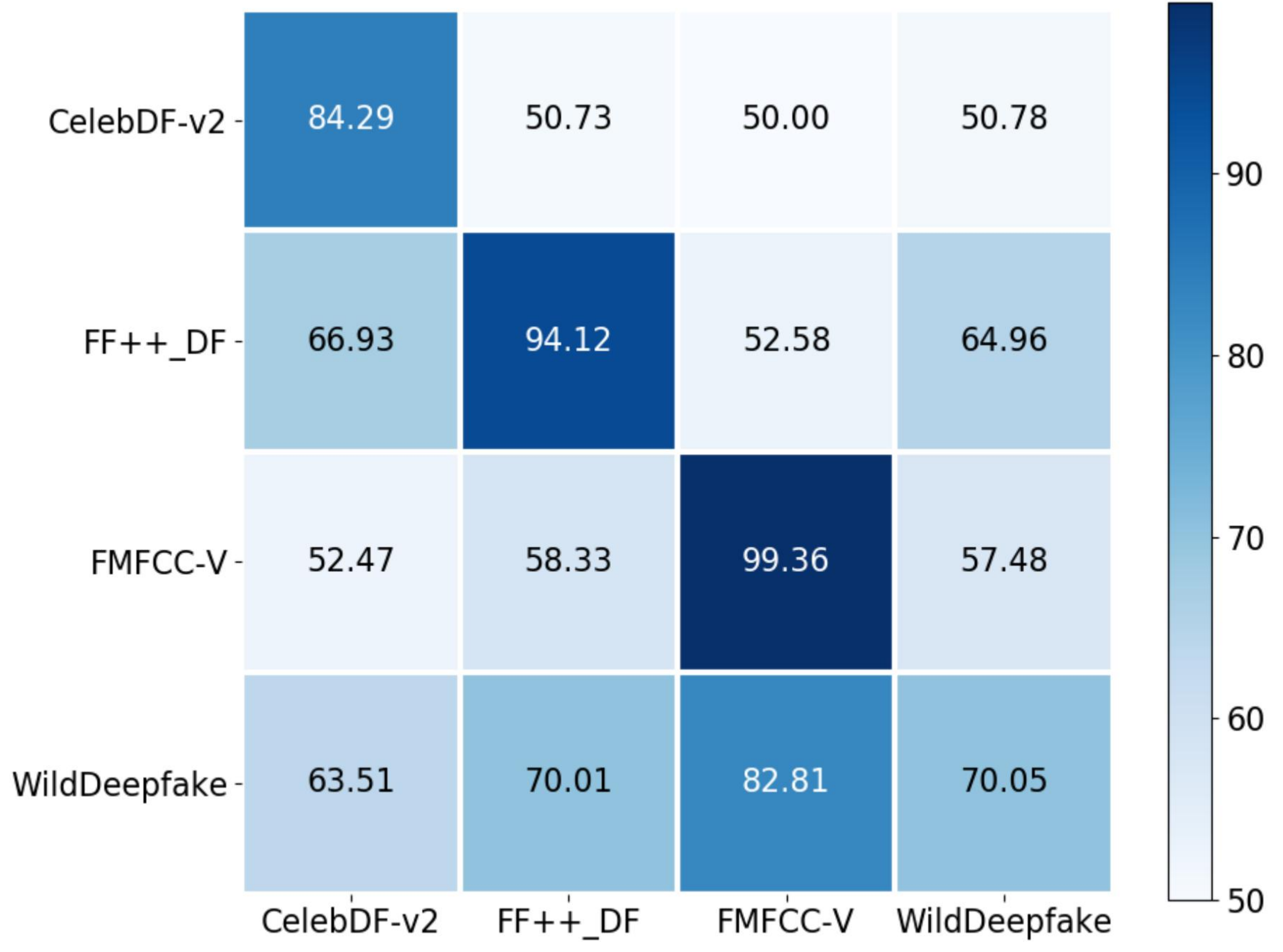}
\caption{Cross-validation results on different training and testing subsets of the Forgery Source Hybrid Dataset without personalized federated learning. The vertical axis represents training data, the horizontal axis represents testing data, and the evaluation index is accuracy.}
\label{fig5}
\end{figure}

As shown in Figure \ref{fig5}, taking the CelebDF-v2 data set as an example, with personalized federated learning, the face forgery detection accuracy is 89.90\%. Without personalized federated learning, the face forgery detection accuracy is 84.29\%. The accuracy rate performance index dropped by 5.61\%. 
Taking the subset of FaceForensics++ (Referred to as FF++\_DF) data set as an example, with personalized federated learning, the face forgery detection accuracy is 94.81\%. Without personalized federated learning, the face forgery detection accuracy is 94.12\%. The accuracy rate performance index dropped by 0.69\%. 
Taking the FMFCC-V data set as an example, with personalized federated learning, the face forgery detection accuracy is 99.60\%. Without personalized federated learning, the face forgery detection accuracy is 99.36\%. The accuracy rate performance index dropped by 0.24\%. 
Taking the WildDeepfake data set as an example, with personalized federated learning, the face forgery detection accuracy is 70.81\%. While without personalized federated learning, the face forgery detection accuracy is 70.05\%. The accuracy rate performance index dropped by 0.76\%. 
It can be seen that without leveraging personalized federated learning, individual clients show poor detection performance due to the lack of shared representation of multiple distributed client models.

\textbf{Ablation study.}
Our methods on the Deepforensics-1.0 in publicly available large-scale datasets
dataset achieved an accuracy of 97.29\%. In the case of centralized data training, our method achieved an identification accuracy of 98.64\%. On the CelebDF-v2 dataset, our method achieved an accuracy of 83.93\% with centralized training. An accuracy of 83.95\% was achieved, clearly proving that the proposed FedPR can further explore more robust face forgery clues. On the Forgery Source Hybrid Dataset, due to complex forgery datasets with diverse types, our method achieved SOTA. The results show that FedPR can not only help improve data security but also maintain strong performance in forgery detection tasks.

To verify the effectiveness of personalized federated learning, we have supplemented additional ablation experiment results as follows.
As shown in Table \ref{tab6}, the result in the first row means that our method removes the personalized forgery representation learning. The accuracy rate on the Deepforensics-1.0 data set is 94.10\%, which is poorer than using our federated face forgery detection learning with a personalized representation method.
Through this experimental result, we verified the effectiveness and advantages of personalized forgery representation. Our approach is able to take advantage of personalized forgery representation enable client models to explore more robust face forgery clues.
The result in the second row is our federated face forgery detection learning with a personalized representation method.
The method has an accuracy of 97.29\% on the Deepforensics-1.0 data set.  
Compared with the method without personalized forgery representation learning, our method improves the accuracy by 3.19\%, it is demonstrated that the personalized features contain attributes that are more suitable for client datasets, further improving the applicability to different types of complex forged datasets.
\begin{table}[t]
	\centering
  \caption{Ablation experimental results of Personalized Forgery Representation on the Deeperforensics-1.0 data set. The evaluation index is face forgery detection accuracy. The best result is displayed in black font and the second place is underlined.}
	\begin{tabular}{ccc}
		\hline	\makecell[c]{Personalized\\Federated Learning}
		  & \makecell[c]{Personalized \\ Forgery Representation} & Accuracy(\%)
		\\
		\hline

~&
\\
\checkmark & - &94.10\\
  - &\checkmark& \textbf{98.64} \\
   \checkmark &\checkmark& \underline{97.29} \\
	\hline
	\end{tabular}

 \label{tab6} 
\end{table}
\section{Conclusions}
In this paper, we explore federated face forgery detection learning with personalized representation, which further explores more robust face forgery clues by combining the shared representation of multiple distributed client models. Furthermore, to improve the applicability of complex forgery datasets with diverse types, the designed personalized forgery representation learning framework can disentangle shared features and personalized features. The personalized features contain more suitable properties of the client data set, further improving the applicability of complex forgery datasets with diverse types. We conducted experiments on public face forgery detection datasets, and the experimental results show that this algorithm can effectively improve the generalization ability of complex forgery datasets with diverse types while protecting privacy. In the future, we will explore extending the personalized federated learning method to cross-modal forgery detection problems, such as combining image and voice information for forgery detection. Through the fusion of cross-modal information and federated learning, the performance and robustness of forgery detection can be further improved. Furthermore, the application of personalized representation in other fields is also worth studying. 

\bibliographystyle{IEEEtran}
\bibliography{ref}
\end{document}